# An Earthworm-Inspired Multi-Mode Underwater Locomotion Robot: Design, Modeling, and Experiments

Hongbin Fang[1,2,3], Zihan He[1,2,3,4], and Jian Xu[1,2,3, §]

[1)] Institute of AI and Robotics, Fudan University, Shanghai 200433, China
[2)] Engineering Research Center of AI & Robotics, Ministry of Education, Fudan University, Shanghai 20043, China
[3)] Shanghai Engineering Research Center of AI & Robotics, Fudan University, Shanghai 200433, China
[4)] School of Aerospace Engineering and Applied Mechanics, Tongji University, Shanghai 200092, China

[§]**Corresponding author:** jian_xu@fudan.edu.cn (J. Xu)

## Abstract

Faced with strong demand for robots working in underwater pipeline environment, a novel underwater multi-model locomotion robot is designed and studied in this research. By mimicking the earthworm's metameric body, the robot is segmented in structure; by synthesizing the earthworm-like peristaltic locomotion mechanism and the propeller-driven swimming mechanism, the robot possesses unique multi-mode locomotion capability. In detail, the in-pipe earthworm-like peristaltic crawling is achieved based on servomotor-driven cords and pre-bent spring-steel belts that work antagonistically, and the three-dimensional underwater swimming is realized by four independently-controlled propellers. With a robot covering made of silicon rubber, the two locomotion modes are tested in underwater environment, through which, the rationality and the effectiveness of the robot design are demonstrated. Aiming at predicting the robotic locomotion performance, mechanical models of the robot are further developed. For the underwater swimming mode, by considering the robot as a spheroid, an equivalent dynamic model is constructed, whose validity is verified via computational fluid dynamics (CFD) simulations; for the in-pipe crawling mode, a classical kinematics model is employed to predict the average locomotion speeds under different gait controls. The outcomes of this research could offer useful design and modeling guidelines for the development of earthworm-like locomotion robot with unique underwater multi-mode locomotion capability.

**Nomenclature:**

| | | | |
|---|---|---|---|
| $m$ | Mass of the robot. | $X_T, Z_T$ | Thrust along the $x$ and $z$ axes. |
| $u, v, w$ | Velocity along the $x$, $y$, and $z$ axis, respectively | $M_T, N_T$ | Moment result of thrust around the $y$ and $z$ axes. |
| $\dot{u}, \dot{v}, \dot{w}$ | Acceleration result of $u$, $v$, and $w$ | $\lambda_{ij}, (i, j = 1, 2, \cdots 6)$ | Fluid inertia force of a body, while moving with unit (Angle) acceleration in an ideal fluid. And 1,2,3 represent the movement along $x$, $y$, and $z$ axes. 4,5,6 represent the rotation along $x$, $y$, and $z$ axes. |
| $a_x, a_y, a_z$ | Acceleration along the $x$, $y$, and $z$ axis, respectively | $\chi_\tau$ $(\chi = X, Y, Z, K, M, N)$ $(\tau = u, v, w, p, q, r)$ | Velocity coefficient, which means that the effect of $\tau$ to $\chi$. |
| $p, q, r$ | Angular velocity about the $x$, $y$, and $z$ axis, respectively | $\phi_{ab}$ $(\phi = X, Y, Z, K, M, N)$ $(a, b = u, v, w, p, q, r, \lvert u \rvert, \lvert v \rvert, \lvert w \rvert, \lvert p \rvert, \lvert q \rvert, \lvert r \rvert)$ | Higher-order velocity coefficient, which means that the effect of $a$ and $b$ to $\phi$. |
| $\dot{p}, \dot{q}, \dot{r}$ | Acceleration of rotation results of $p$, $q$, and $r$. | $L_L, L_R, L_F, L_B$ | Distances between the robot's mass center to left, right, front and back propellers. |
| $Jx, Jy, Jz$ | Moment of inertia about the $x$, $y$, and $z$ axis, respectively | $V_\xi, V_\eta, V_\zeta$ | Velocity of the robot on the global coordinate system. |
| | | $\alpha$ | Angle between the x-axis and $\xi$ axis. |

# 1. Introduction

Earthworm, a kind of soft creature that lives in the damp cave of underground, can move in clay, detritus, ruins, grass. For these locomotion abilities and concealment abilities, many researchers try to study earthworm-like robots[1,2], which can be used for ruins search, pipeline inspection, and battle reconnaissance. All these abilities owe to the earthworm's morphology characteristics[3–5], which contain three main factors. First, the multi-segment structure: the earthworms contain a good deal of segments; the segment separated by the cuticle is an independent structure to control the segment's morphological characteristics. Second, antagonistic muscle: the longitudinal muscle and the circular muscle are a pair of antagonistic muscles. As the longitudinal muscle contracted, the circular muscle is released, the axial length is decreased, and the radial length is increased. On the other hand, as the circular muscle contracted, the longitudinal muscle is released, the axial length is increased, and the radial length is decreased. Third, setae: the anisotropic friction is crucial for the body to move. Setae can change the coefficient of friction, which makes the backward friction factor is bigger than the forward friction factor. With the setae, the earthworm can move forward. Fourth, the retrograde peristalsis wave: the multi-segment can coordinate to contract or relax the longitudinal muscle and circular muscle, which can generate the peristalsis wave. With the peristalsis wave spread back, the earthworm can be driven forward. All these morphology characteristics and locomotion mechanisms make the earthworm a unique creature and impel researchers to study earthworm-like robots.

The wheeled and legged robot can't move efficiently in ruins, because the wheels and legs can be locked by macadam. However, due to the continuous appearance, the earthworm-like robot can effectively move in ruins, pipelines[6], and battlegrounds. The crawling robots have a slow speed of movement which can improve the accuracy of detection. However, the robots are not good at the rapid movement. For example, as the robots are used for underwater pipe inspection, the robot needs carriers to transmit the robot to the workplace to improve efficiency[7]. To improve the locomotion ability and not lost the earthworm's characteristics, an earthworm inspired swimming-crawling multi-mode locomotion robot is designed and experimented with. We wish that the multi-mode robot can be used to explore underwater resources and inspect the underwater pipeline.

In our daily life, the robots are defined into multi-mode robots by two factors, the environment, and personal factors. For the first classification, the robot can move in two or more environments in water, land, and sky. For example, the aquatic-terricolous multi-mode robot can swim in the water and walk or roll on land[8–11], the aquatic-aerial multi-mode robot can swim in the water and fly in the sky[12–18], the terricolous-aerial multi-mode robot can walk, jump, crawl, or roll on land and fly in the sky[19–28]. For the second classification, the robot can change the personal locomotion mode into two or more locomotion modes in the same environment. For example, the walking-rolling robot can drive the legs to walk and drive the wheels to roll[29–34], the walking-creeping robot can change personal structure to walk and creep[35–38], the spray water and swimming multi-mode robot not only can use feet to swim but also can spray water to drive the body to move[39]. All in all, the multi-mode robots can enhance the robot's adaptive capacity to the environment and maximize the value of the robot.

In this study, a multi-mode robot is designed and experimented, and the robot can change the personal locomotion mode into swimming mode in the water and crawling mode in the underwater pipe. The robot mainly contains two modules, the swimming module, and the crawling module. To improve movement efficiency, the robot swimming module uses the propellers and DC motors as the driver, which consists of two horizontal propellers to drive the robot move on the horizontal plane, and two vertical propellers to drive the robot moves on the vertical plane. On the other hand, the crawling module is constituted of six peristalsis segments, which use the servo motor as the driver. To prevent water damage the servo motor, several moulds are designed to fabricate silicone skins. For the control mode, the computer system generates signals and sents to the Arduino control board. After being handled, the signals are sent to the servo motor control board to drive the six servo motors, and the DC motor driver to drive the four propellers. To predict the velocity and location of the robot, the dynamic equations are established by drawing from the submarine's dynamic equations. With the equations, the robot's locomotion on the horizontal plane and vertical plane can be predicted. Especially, the robot can move with circular locomotion on the horizontal plane, linear locomotion on the horizontal and vertical plane. Furthermore, the CFD software is used to simulate the three basic locomotions and compared them with ODE45 numerical simulation based on the dynamic equations.

This article is organized as follows: The robot design and prototype are presented in section 2. The robot control and locomotion test are presented in section 3. The modeling and locomotion analysis are presented in section 4. Finally, the discussion and conclusion are presented in section 5.

## 2. Robot design and prototype

In this study, a swimming-crawling multi-mode locomotion robot is designed. The earthworm's biological structures are studied as the peristaltic module. The propeller is a kind of mature and reliable part, so it is used as the driver for the swimming module in this robot. By assembling the two kinds of modules a multi-mode robot is designed and prototyped.

### 2.1. Structure design

To achieve swimming and acrawl modes a multi-mode robot is designed(figure 1(a)). In this design, the parts of the swimming module and the parts of the peristalsis module are alternated to make the mass more uniform. To drive the robot to arise and down in the water, two vertical swimming module parts(figure 1(b)) are designed. The two part's propellers are twain, which can offset the torque to the robot body they produced. The intermediate part is the horizontal swimming module part, which can drive the robot moving on a plan by defining different rotational speeds for left and right propellers. The robot can swim in capacious water, however, the left and right propellers may hinder the robot to enter to some pipes, so the two propellers need to behave the ability to contract (figure 1(c))and expand(figure 1(d)). On the other hand, swimming is more efficient than peristalsis, so the robot can crawl through the environment, when the left and right propellers can expand the robot can swim at a high speed. To achieve the function, two parallel four-bar linkages are driven by a waterproof servo motor(figure 1(d)).

Three features of the earthworm structure are longitudinal muscle, circular muscle, and the metameric segmentation of the whole earthworm body. Figure 1(e) shows the cross-section of the earthworm. By imitating the earthworm's main features, a robot segment is designed, which uses the servo-motor-driven cords to imitate longitudinal muscle, spring-steel as circular muscle, hand plate, and end plate as the septums between segments. Figure 1(f) shows the relaxed segment, by driving the servo-motor-driven cords the segment can be compressed on contracted condition(figure 1(g)). The robot has two peristalsis elements, peristalsis elements A that contains one peristalsis element, and peristalsis elements B that contains two peristalsis elements. When the swimming parts and peristalsis parts are connected, the robot can swim in the water and crawl in the pipe. To reduce the Cross-section resistance, which is the main resistance, two cusps are designed on the head and end of the robot.

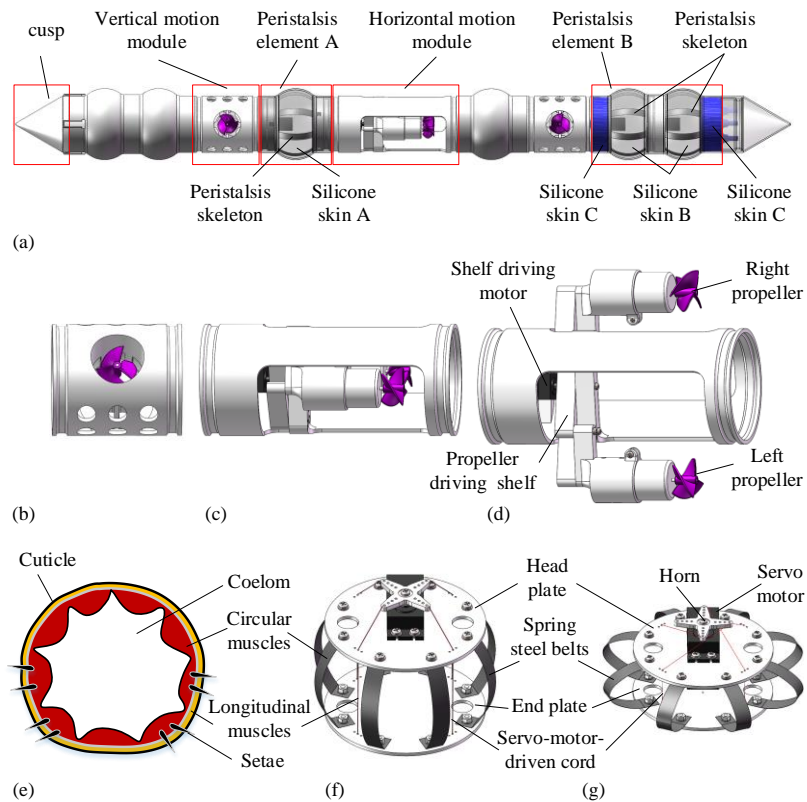

Figure 1. multi-mode robot structure design. (a) The integral structure design; (b) Vertical locomotion structure for swimming; (c) Vertical locomotion structure for swimming on contracted condition; (d) Vertical locomotion structure for swimming on expanded condition; (e) Schematic illustrations of the cross-section of an earthworm; (f) CAD design of a single segment of the earthworm-like robot on expanded condition; (g) CAD design of a single segment of the earthworm-like robot on contracted condition.

## 2.2. Skin design

Peristalsis parts can't immerse in water for a long time, to change it, the waterproof skins need to be designed. The fabrication processes are shown in figure 2(a). First of all, a bottom mould, an inner mould, and two outer moulds need to be designed and printed by 3D printer. Then assemble the modules. On the other hand, silicone(smooth on ecoflex 00-30) part A and part B mixed in equal proportion. After being stirred uniformly, the mixed silicone is poured into the moulds. Put moulds in the vacuum bin for about 20 minutes, then take it out. After the silicone is solidified, the silicone skin can be taken out. With similar processes, three kinds of silicone skins(figure2.(b), figure2.(c), figure2.(d)) can be made. Silicone part A can enclose two robot segments, part B can enclose one segment, and part C cover each side of the segment. Then the robot segments can be wrapped up by the three kinds of silicone skin. To through wires, three tentacles are designed at the end of silicone C. The tentacles can increase airtightness between silicon skin and wires to prevent water come in.

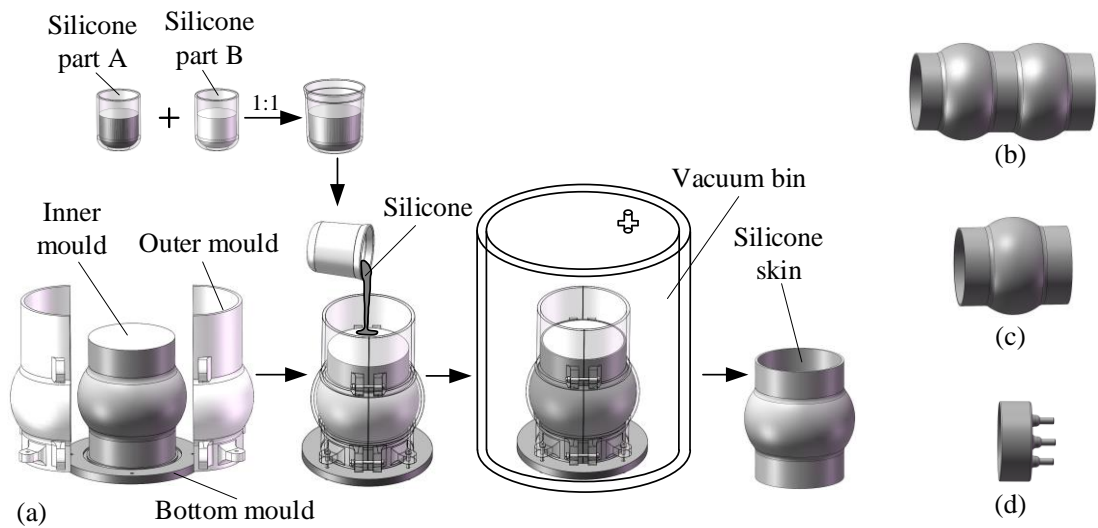

Figure 2. Mold processes and three kinds of silicone skins. (a) Mold processes;(b) Silicone skin A; (c) Silicone skin B; (d) Silicone skin C.

## 2.3. Prototype design

The figure3 shows the prototype of the assembled multi-mode robot. The total length of the robot is 1340mm, and the diameter of the robot in the relaxed condition is 105mm. when the left and right propellers expand the length of the two propellers' axes is 145mm. The weight of the robot is 5.542kg, which nearly equal to the weight of the same volume of water, so the robot can suspend in water.

Each peristalsis segment assembled by two acrylic plates which are 2mm thick; eight spring-steel belts which are 80mm in length, 10mm in width, and 0.1mm in thickness; a servomotor with the torque of 21.8kg·cm, whose weight is 58g (JX DC5821LV); and the servomotor as the driver to pull Berkley Fireline Braid (Model:TBFS65-22, 0.381mm in diameter, made out of Dyneema, 65lb (29.5kg) pound test) as the servomotor-driven cord. To prevent water, the segments are contained by the skins which are made of silicon (Smooth-on Ecoflex 0030).

The swimming module uses 2 DC-motor (Model: ipiggy, 015-hgb) as the horizontal drivers and the same 2 DC-motor as the vertical drivers. To contract and expand the left and right propellers the two parallel four-bar linkages are driven by a waterproof servo motor (JX DC5821LV, a torque of 21.8kg·cm, 58g weight). To let the robot suspend in water, the body of the swimming module is made of stainless steel (316L, 7.98g/cm³) by 3D print.

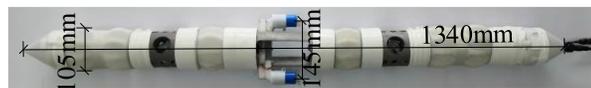

Figure3. The prototype of the multi-mode robot and it's dimensions.

## 3. Robot control and locomotion test

In this section, the robot control system and the results of theoretical gait analysis and swimming locomotion analysis are studied based on the swimming-crawling multi-mode locomotion robot.

### 3.1. Controller design

Robot control is an important part of the locomotion robot. As shown in figure4 the control system includes two main parts the aquatic locomotion control system and the in-pipe locomotion control system. When the computer control system generates a signal and sends it to the Arduino control board, the board judges the signal belong to which module and sends it to that control board. The aquatic locomotion control system includes two DC motor drivers to control the front propeller, back propeller, left propeller, and right propeller. And the Arduino control board direct controls a servo motor to expand or contract the left and right propellers. On the other hand, the in-pipe locomotion control system controls 6 servo motors by servo motor control board to carry out different gaits.

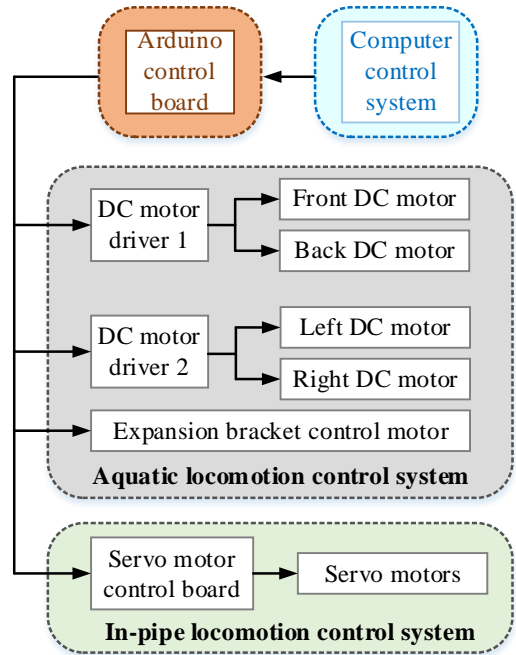

Figure4. Flow chart of the control system.

## 3.2. Test environment

To test the locomotion characteristics, a swimming pool (Bestway, 2.59*1.70*0.61m) is loaded with 0.4 meters of water. Three waterproof cameras(GoPro HERO8 Black 4K) are respectively set in the front, side, and upward side to record the locomotion. The computer control system transmits signals by signal line to control the robot and the robot gets power (DC, 7.0V, 5.5A) by power line from the power suppliances (KPS3010D, 0-30.00V, 0-10.00A, 300W).

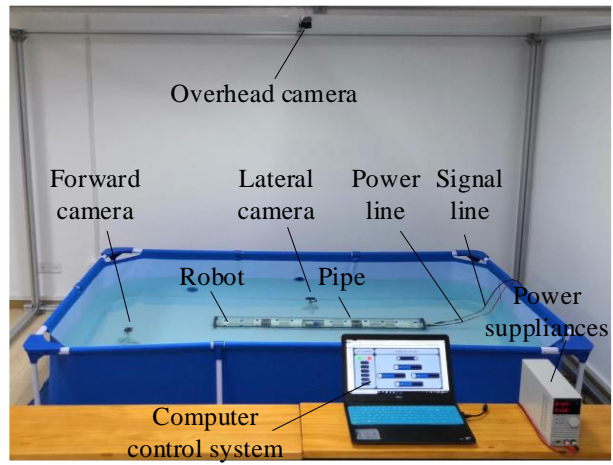

Figure5. The robot test environment.

### 3.3. In-pipe locomotion

To show the robot's locomotion ability in the underwater pipe, the multi-mode robot is tested in the horizontal underwater pipe(1.8m length, 129mm inner diameter) with the gait $n_A = 1, n_R = 1, k = 1$ and actuation angle 145 degrees. And the robot needs per second to change the state from one to the next.

To clearly study the robot's locomotion characteristics and retrograde peristalsis wave, some video frames are picked out. As shown in figure 6(b), every moment the robot has one segment to anchor with the pipe and as time goes on the anchored segment moves back. On the other hand, the period of the present gait is 6 seconds and the robot moves 10 periods from $7^{th}$ seconds to $67^{th}$ seconds. Figure 6(a) shows the displacement-time history, which shows that in the 10 periods the displacement of the robot is 0.428m. So the robot moves at 7.13mm/s with the gait $n_A = 1, n_R = 1, k = 1$.

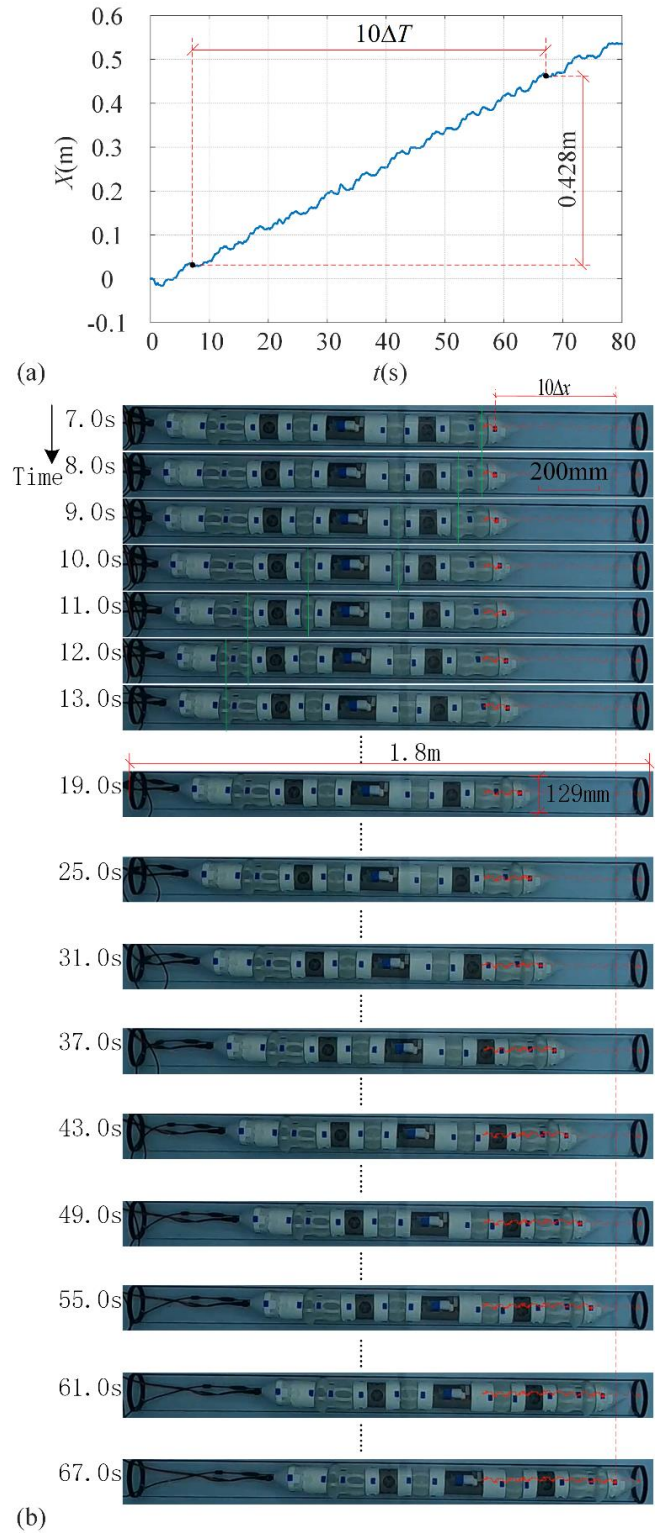

Figure6. Horizontal test in peristalsis mode for the multi-mode robot in the underwater pipe with the gait $n_A = 1, n_R = 1, k = 1$ and actuation angle 145 degrees. (a) the displacement-time history of the robot. (b) the frames pick out from the experiment video.

## 3.4. Aquatic locomotion

To verify the multi-mode robot's swimming ability, the three basic locomotion is studied. First, figure 7 shows the Forward and backward tests in swimming mode for the multi-mode robot in the water. The mark is made in the robot's head. By tracing the mark, the trajectory can be shown in figure 7(a). Six points(A, B, C, D, E, F) are selected in the trajectory, and the corresponding frames are picked out as shown in figure 7(b). Figure 7(a) shows that the robot moves toward 1.15m, and hit against the wall of the swimming pool and make it a little oblique. When the robot comes back, it deviates the original position 0.10m.

Figure 8 shows the sinking and rise tests in swimming mode for the multi-mode robot in the water. With the same processes, we select six different moments and pick out the frames from the experiment video as shown in figure 8(b). Figure 8(a) shows that when the multi-mode robot sinks to 0.229m and rises, the robot deviates from the original position of 0.077m.

The robot has the ability to swimming on a plan. To verify the robot's ability, we control the robot to make left circle locomotion as shown in figure 9. As the same, we select six different moments and pick out the frames from the experiment video as shown in figure 9(b). The robot's locomotion trajectory is shown in figure 9(a), with the trajectory approximate circle can be fitted. Before point C, the robot is unstable, so it doesn't fit very well. However, behind point C, the locomotion trajectory and approximate circle can be fitted very well, and the radius of the approximate circle is 2.980m.

The mode changeability for the multi-mode robot is significant. When the multi-mode robot crawls from a pipe, the robot needs to change the locomotion from peristalsis mode to swimming mode. Figure10 shows the multi-mode robot crawl from a pipe. When the propeller driving shelf moves out of the pipe, the left and right propellers are expanded by the driving motor. Then the left and right propeller driving the robot to move out of the pipe. Figure 10(a) shows the multi-mode robot in-pipe locomotion trajectory and swimming locomotion trajectory. Some frames are picked out from the experiment video, while the robot crawls in the pipe at the moment A, B, and C, the left and right propellers are expanded at the moment D, and swimming forward until moment E.

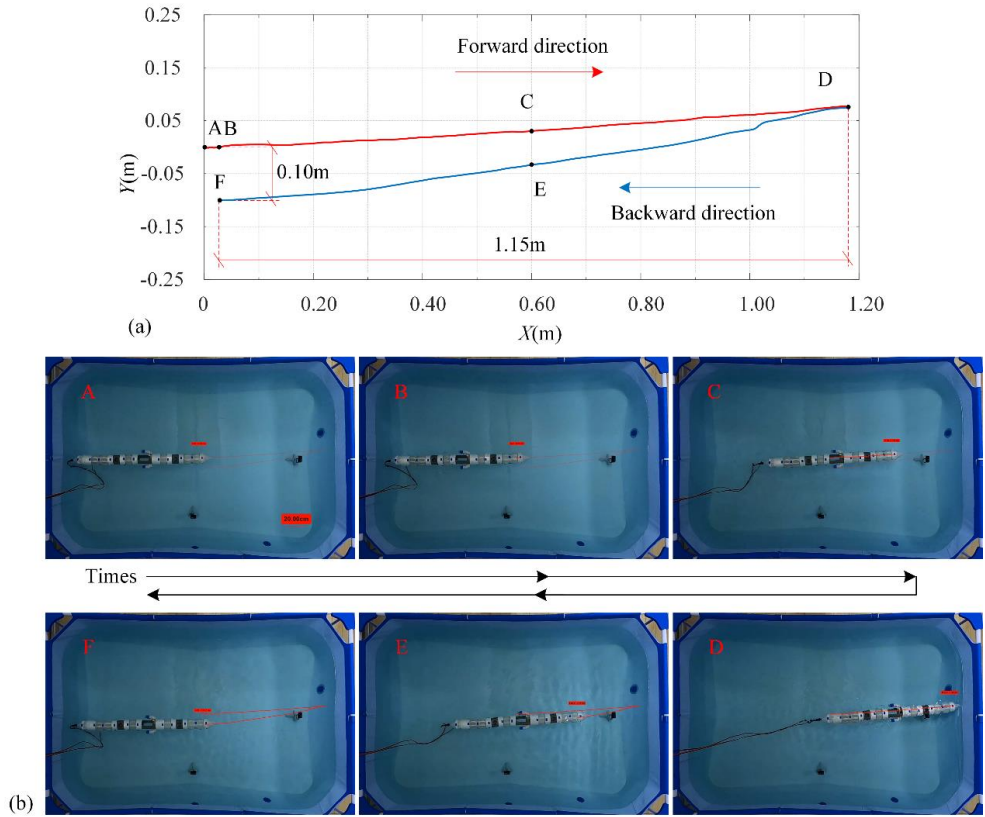

Figure7. Forward and backward tests in swimming mode for the multi-mode robot in the water. (a) the locomotion trajectory of the robot. (b) the frames pick out from the experiment video.

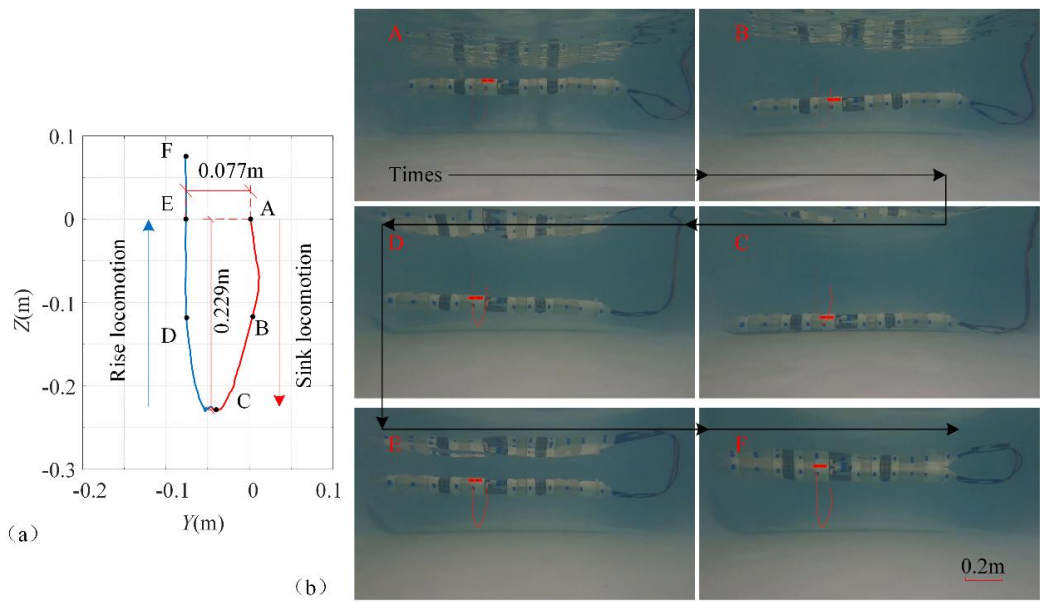

Figure8. Sinking and rise tests in swimming mode for the multi-mode robot in the water. (a) the locomotion trajectory of the robot. (b) the frames pick out from the experiment video.

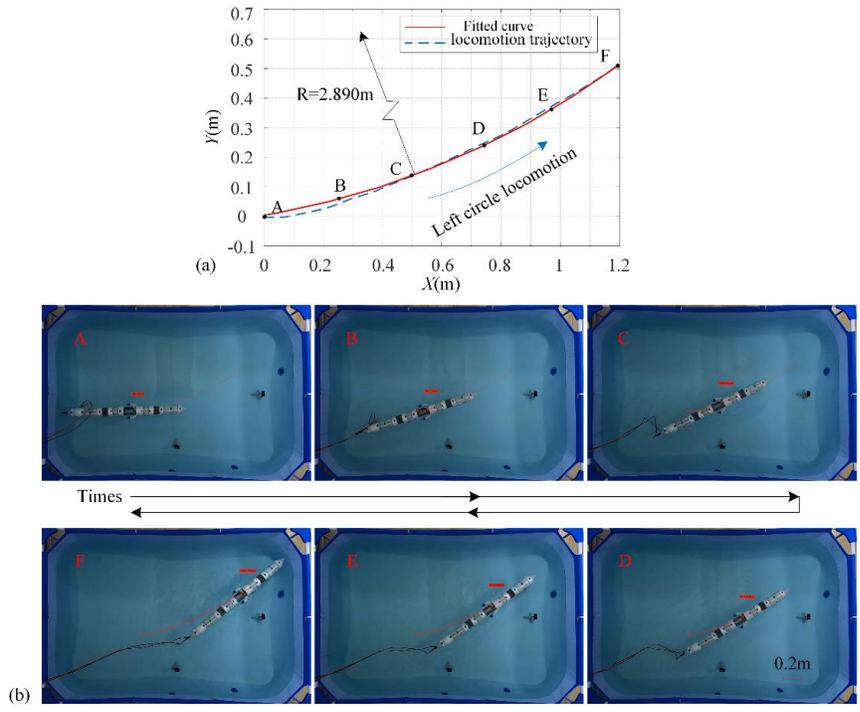

Figure9. Left circle test in swimming mode for the multi-mode robot in the water. (a) the locomotion trajectory of the robot. (b) the frames pick out from the experiment video.

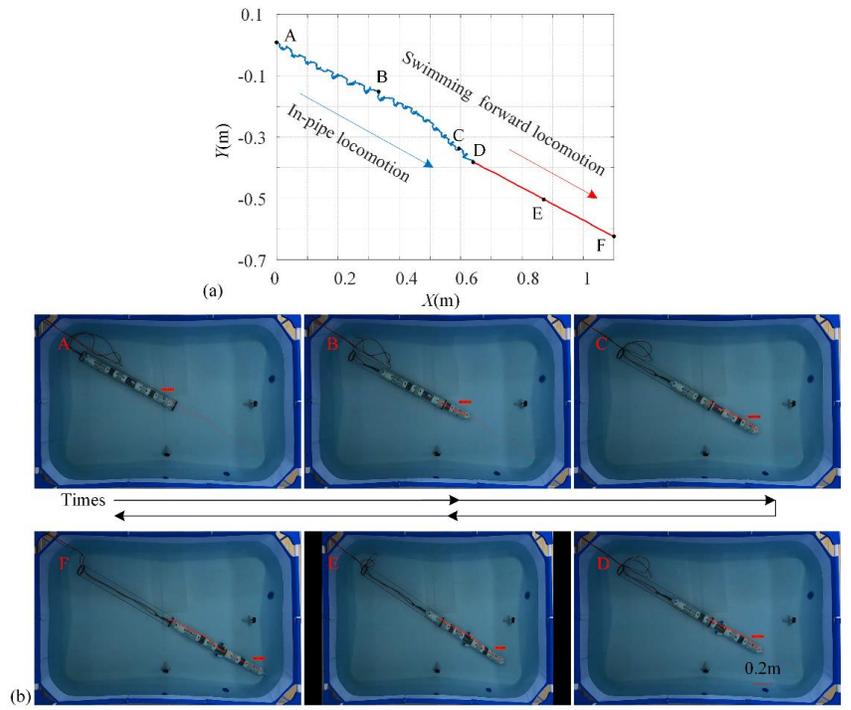

Figure10. Locomotion mode change test from peristalsis mode to swimming mode for the multi-mode robot in the water. (a) the locomotion trajectory of the robot. (b) the frames pick out from the experiment video.

### 3.5. Discussion

In this section, the multi-mode robot's effective locomotion ability is verified. First of all, the robot's control system is designed to control the robot to move in various states. Second, the test environment is established, which includes the three GoPro waterproof cameras, swimming pool, and acrylic pipe. Third, we test the multi-mode robot in the pipe with peristalsis mode; swimming mode which includes forward and backward test, sinking and raise test, left circle test; and then we test the multi-mode robot's mode changeability that the robot crawls from a pipe and changes the locomotion mode to swimming mode to continue moving forward.

## 4. Modeling and locomotion analysis

In this section, the robot's control model and locomotion characteristics are analyzed. First, based on the submarine's control theories, we establish the robot's control model in the swimming mode. Second, to confirm the model, we use CFD software to simulate the three basic locomotions. Then we compare the control model's results and CFD simulation results. Furthermore, we analyze the relationship between velocities, diameters, and different propeller forces. At last, based on the kinematic model we analyze the peristalsis locomotion ability in the pipe for the multi-mode robot.

### 4.1. Dynamic model for aquatic locomotion

To predict the locomotion of the multi-mode robot, we reference the dynamic model of the submarine[40]. A reliable dynamic model can not only reduce the research and development cost but also shorten the research and development cycle. To simplify the model appropriately, some assumptions are as follows:

1. When the robot is in the water, the buoyancy always equal to gravity and the center of gravity coincides with the center of gravity of buoyancy, which means that we don't need to consider buoyancy and gravity.

2. On the submarine's fluid theory, the robot is regarded as a spheroid.

3. The mass center and origin of the local coordinate system are at the same point.

4. The three axes of the local coordinate system always point to the robot's front, right, and underneath.

As shown in figure 11, $E-\xi\eta\zeta$ is the global coordinate system. The original point $E$ is any point on the horizontal plane. The $E\xi$ axis is on the horizontal plane, and it also is the main direction of navigation. The $E\eta$ axis is on the horizontal plane too, and it also is the side navigation direction. The $E\zeta$ axis is perpendicular to the plane $E\xi\eta$, and also follows the right-hand rule. $G-xyz$ is the local coordinate system. $G$ is the center of the mass for the robot. The positive directions of $Gx, Gy$ and $Gz$ respectively point to the front, the right, and the underneath of the robot. The axes $Gx, Gy, Gz$ are the principal axes of inertia for the robot. The $K$, $M$, and $N$ are the positive directions of torques. The $X$, $Y$, and $Z$ are external forces along the three axes.

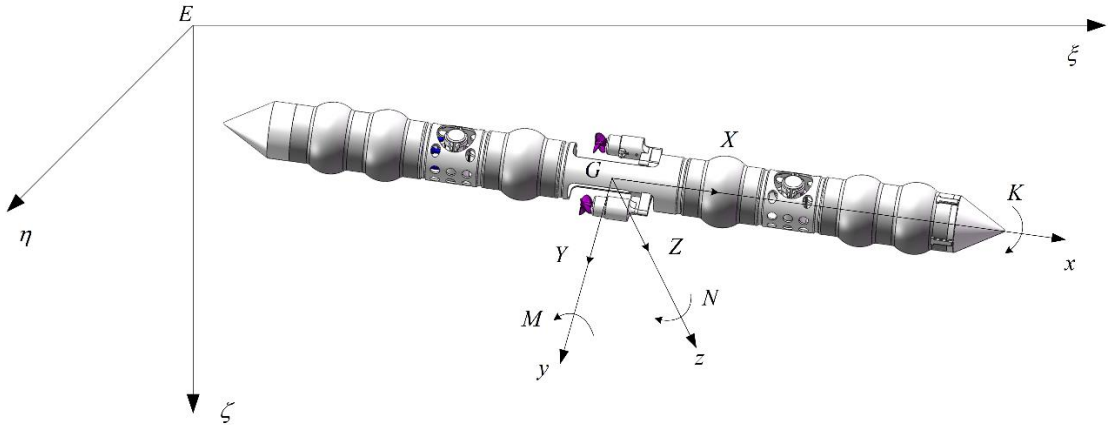

Figure11.the robot in the global and local coordinate systems

With the momentum theorem and Euler's dynamic equation, the submarine has six degrees of freedom in the local coordinate system, which includes translations and rotations in three directions can be shown

$$\begin{cases} ma_x = X_I + X_{viscous} + X_T \\ ma_y = Y_I + Y_{viscous} \\ ma_z = Z_I + Z_{viscous} + Z_T \\ J_x = K_I + K_{viscous} \\ J_y = M_I + M_{viscous} + M_T \\ J_z = N_I + N_{viscous} + N_T \end{cases} \quad (1)$$

In this study, we only research three typical movements in the horizontal plane and vertical plane, including circular movement in the horizontal plane, forward and backward in the horizontal line, arise and down in the vertical line. The dynamic equation on the horizontal plane can be shown

$$\begin{cases} ma_x = X_I + X_{viscous} + X_T \\ ma_y = Y_I + Y_{viscous} \\ J_z = N_I + N_{viscous} + N_T \end{cases} \quad (2)$$

where

$$\begin{cases} ma_x = m(\dot{u} - rv) \\ ma_y = m(\dot{v} + ru) \\ J_z = I_z \dot{r} \\ X_I = -\lambda_{11}\dot{u} - \lambda_{33}wq + \lambda_{22}vr \\ Y_I = -\lambda_{22}\dot{v} - \lambda_{11}ur + \lambda_{33}wp \\ N_I = -\lambda_{66}\dot{r} + (\lambda_{11} - \lambda_{22})uv + (\lambda_{44} - \lambda_{55})pq \\ X_{viscous} = -\left(X_{uu}u^2 + X_{vv}v^2 + X_{rr}r^2 + X_{vr}vr\right) \\ Y_{viscous} = -\left(Y_v v + Y_r r + Y_{v|v|}v|v| + Y_{r|r|}r|r| + Y_{v|r|}v|r|\right) \\ N_{viscous} = -\left(N_v v + N_r r + N_{v|v|}v|v| + N_{r|r|}r|r| + N_{v|r|}v|r|\right) \\ X_T = F_L + F_R \\ N_T = F_L * L_L - F_R * L_R \end{cases} \quad (3)$$

The control equation for circular movement in the horizontal plane can be shown

$$\begin{cases} m\dot{u} = -\left(\lambda_{11}\dot{u} + X_{uu}u^2 + X_{rr}r^2\right) + F_L + F_R \\ I_z \dot{r} = -\left(\lambda_{66}\dot{r} + N_r r + N_{r|r|}r|r|\right) + F_L * L_L - F_R * L_R \end{cases} \quad (4)$$

Eq.(4) shows the control equation in the local coordinate system, while the equation translates into the global coordinate system can be shown

$$\begin{cases} (m+\lambda_{11})\dfrac{d(V_\xi \sec\alpha)}{dt} = -\left(X_{uu}(V_\xi \sec\alpha)^2 + X_{rr}\left(\dfrac{d\alpha}{dt}\right)^2\right) + F_L + F_R \\ (m+\lambda_{11})\dfrac{d(V_\eta \csc\alpha)}{dt} = -\left(X_{uu}(V_\eta \csc\alpha)^2 + X_{rr}\left(\dfrac{d\alpha}{dt}\right)^2\right) + F_L + F_R \\ (I_z+\lambda_{66})\dfrac{d^2\alpha}{dt^2} = -\left(N_r \dfrac{d\alpha}{dt} + N_{r|r|}\left(\dfrac{d\alpha}{dt}\right)^2\right) + F_L * L_L - F_R * L_R \end{cases} \quad (5)$$

The control equation for forward and backward in the horizontal line can be shown

$$(m+\lambda_{11})\dfrac{dV_\xi}{dt} = -X_{uu}V_\xi^2 + F_L + F_R \quad (6)$$

The dynamic equation on the vertical plane can be shown

$$\begin{cases} ma_x = X_I + X_{viscous} + X_T \\ ma_z = Z_I + Z_{viscous} + Z_T \\ J_y = M_I + M_{viscous} + M_T \end{cases} \quad (7)$$

where

$$\begin{cases} ma_x = m(\dot{u} + qw) \\ ma_z = m(\dot{w} - qu) \\ J_y = I_y \dot{q} \\ X_I = -\lambda_{11}\dot{u} - \lambda_{33}wq + \lambda_{22}vr \\ Z_I = -\lambda_{33}\dot{w} + \lambda_{11}uq - \lambda_{22}vp \\ M_I = -\lambda_{55}\dot{q} + (\lambda_{33} - \lambda_{11})uw + (\lambda_{66} - \lambda_{44})pr \\ X_{viscous} = -\left(X_{uu}u^2 + X_{vv}v^2 + X_{rr}r^2 + X_{vr}vr\right) \\ Z_{viscous} = -\left(Z_w w + Z_{|w|}|w| + Z_q q + Z_{w|w|}w|w| + Z_{ww}w^2 + Z_{w|q|}w|q| + Z_{q|q|}q|q|\right) \\ M_{viscous} = -\left(M_w w + M_{|w|}|w| + M_q q + M_{w|w|}w|w| + M_{ww}w^2 + M_{w|q|}w|q| + M_{q|q|}q|q|\right) \\ X_T = F_L + F_R \\ Z_T = F_F + F_B \\ M_T = -F_F * L_F + F_B * L_B \end{cases} \quad (8)$$

The dynamic equation for arising and down in the vertical line can be shown

$$(m + \lambda_{33})\dot{w} = -\left(Z_w w + Z_{w|w|}w|w|\right) + (F_F + F_B) \quad (9)$$

Eq.(9) shows the control equation in the local coordinate system, while the equation translates into the global coordinate system can be shown

$$(m + \lambda_{33})\frac{dV_\varsigma}{dt} = -\left(Z_w V_\varsigma + Z_{w|w|}V_\varsigma |V_\varsigma|\right) + (F_F + F_B) \quad (10)$$

With Eq.(5), we can calculate some characteristics such as the location and velocity of the robot. Eq.(6) can predict the characteristics while the robot moves forward and backward. Eq.(10) can predict the characteristics while the robot moves arise and down. With the three basic locomotion, we can verify the reliability of the dynamic equation.

## 4.2. Three basic aquatic locomotion

To confirm the robot's control theory, we verify the three basic locomotions with CFD software simulation. We import the CAD module into CFD simulation software, define the propeller's force, and simulate based on CFD theory. On the other hand, by defining the propeller's force, we calculate the robot's locomotion trajectories and velocities with the Eq(5), Eq(6), and Eq(10). The moment of inertia with three axes are $I_x = 0.007239 \text{kg} \cdot \text{m}^2$, $I_y = 0.684939 \text{kg} \cdot \text{m}^2$, $I_z = 0.685918 \text{kg} \cdot \text{m}^2$, the mass of the total robot is $m = 5.542 \text{kg}$, the gravity acceleration is $g = 9.8 \text{kg} \cdot \text{m/s}^2$, and by defining the propellers with different force to drive the robot to move in a different state.

As shown in figure 12, the comparison between CFD software simulation and numerical calculate with ode45. We define the robot's left propeller with 10N and right propeller with -5N, the robot's locomotion trajectories can be seen in Figure 12(a). The robot star from the initial position and convergent to a circle. When the robot in a stable condition, the diameter of CFD simulation is 1.756m, and the diameter of ODE45 calculated is 1.733m, from that we can find the two methods have the same result, and the control equation can predict the robot's locomotion in some extent. Figure 12(b) shows the tangential velocities when the robot moves in circular locomotion. When the robot is in a steady-state the CFD software simulation velocity is 0.855m/s, and the ODE45 numerical calculate velocity is 1.079m/s. The CFD simulation result is a little small than the ODE45 result. On the other hand, the trajectories and the rate of convergence shown in figure 12(a) are a little different, all these reasons are resulted from that the control theory the robot is simplified into a cylinder and the left and right propellers' hydrodynamic force and torque are not considered, which results that the ODE45 calculate results have a little difference with CFD software simulation.

Figure 12(c) reveals that the velocity of the robot in forward locomotion. The left propeller force and right propeller force are equal, and the resultant force is 0.6N. The CFD simulation result is 4.61m/s, and the ODE45 calculation result is 0.374m/s, the error is not too great. Figure 12(d) demonstrates the velocity of the robot in sinking locomotion. The front propeller force and back propeller force are equal, and the resultant force is 2N. The CFD simulation result is 0.251m/s, and the ODE45 calculation result is 0.234m/s, the error is not too great too.

All in all, with the three basic locomotion on CFD software simulation and ODE45 numerical calculation, the control equation can be verified to predict the robot's locomotion abilities.

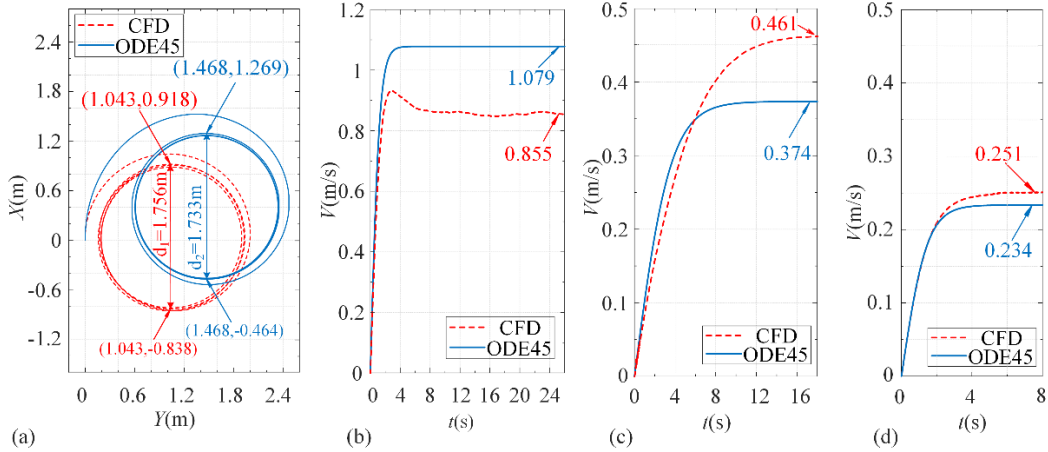

Figure12. comparison between CFD software simulation and numerical calculation. (a) the locomotion trajectories of the robot in circular locomotion. (b) the velocity of the robot in circular locomotion. (c) the velocity of the robot in forward locomotion. (d) the velocity of the robot in sinking locomotion.

### 4.3. Locomotion analysis

As mentioned above, the robot's control model is verified effective. In this section based on the control model, further analysis is studied on the three basic locomotion. Figure 13(a), the full line shows the relationship between velocities and the coupled propeller force for equal left and right propeller force, and the dashed line shows the relationship between velocities and the coupled propeller force for equal front and back propeller forces. With the same coupled force the robot horizontal line direction is always fast that the vertical line direction, the reason is that the resistances on the cross-section is small than on the longitudinal section.

When the robot moves in the horizontal plane if the left propeller force and right propeller force are not equal, the robot will take circular locomotion. Figure 13(b) reveals the relationship between tangential velocities and coupled left and right propeller forces. We can find that if the left propeller force equal negative right propeller force the velocity always equal 0, which means that the robot rotates about the original point. In figure 13(a) the top left corner means that the robot makes the left-circle locomotion, and the bottom right corner means that the robot makes the right-circle locomotion. Figure 13(c) shows the relationship between diameter that after being taken logarithm and coupled left and right propeller forces. which mean $LD = \log_{10}(Diameter)$.

The figure shows that if the left propeller force equal negative right propeller force the diameter always equals 0, which means that the robot rotates about the original point and $LD$ equals infinitesimal. And while the left propeller force nearly equal right propeller force, the diameter will tend to infinity and LD equals infinite. So in figure 13(c) contains two special regions the infinite region and the infinitesimal region.

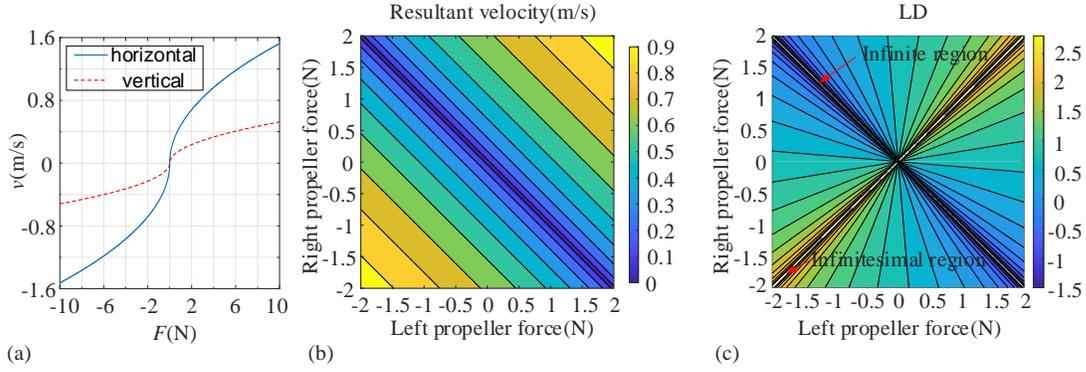

Figure 13. Locomotion analysis of three basic locomotion. (a) the relation between the coupled force and horizontal velocity and vertical velocity on linear locomotion. (b) the resultant velocity of the robot with coupled left propeller force and right propeller force. (c) the locomotion diameter after taking the logarithm of the robot with coupled left propeller force and right propeller force.

### 4.4. Kinematic model for in-pipe locomotion

The kinematic model for the in-pipe robot has been in-depth studied[4]. In his research, the velocity can be obtained with displacement divided by time as

$$\bar{V} = \frac{X}{T} = \frac{\mu[N - k(n_A + n_R)]\Delta l}{\mu(N/n_R)\Delta t} = \frac{N - k(n_A + n_R)}{N/n_R} \frac{\Delta l}{\Delta t} \quad (11)$$

➢ $N$ : the total number of the segments for the robot;

➢ $k$ : the number of driving modules;

➢ $n_A$ : the number of anchoring segments in a driving module;

➢ $n_R$ : the number of relaxing/contracting segments in a driving module;

➢ $\Delta l$ : the deformation of a single segment in the axial direction;

➢ $\Delta t$ : the time of one actuation;

➢ $\mu$ : the coefficient.

In this research, we only study gaits with one driving module and $N = 6, k = 1, \Delta l = 2.773, \Delta t = 1$. Table 1. shows the gait parameters. With the gait parameters, the average speed in each gait can be calculated and shown in figure 14. From which we can find that the robot with one anchoring segment and two relaxing/contracting segments can reach the maximal speed of 2.77cm/s, and four anchoring segments and one relaxing/contracting segment can reach the minimum speed of 0.46cm/s.

Table 1. gait parameters for the multi-mode robot.

| Gait No. | $k$ | $n_A$ | $n_R$ |
|---|---|---|---|
| 1 | 1 | 1 | 1 |
| 2 | 1 | 2 | 1 |
| 3 | 1 | 3 | 1 |
| 4 | 1 | 4 | 1 |
| 5 | 1 | 1 | 2 |
| 6 | 1 | 2 | 2 |

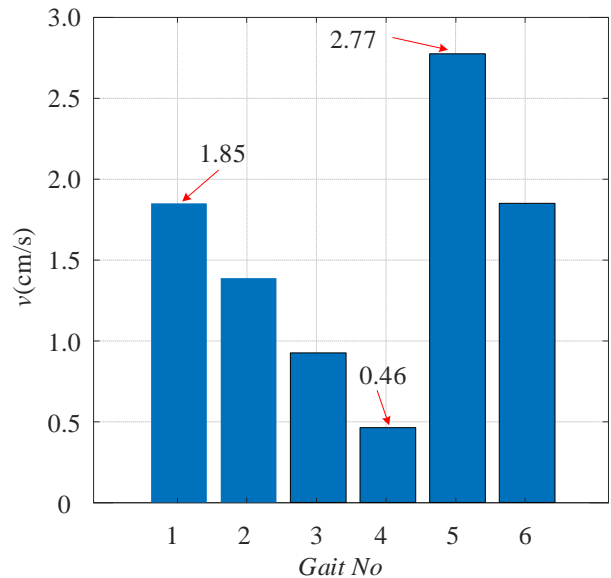

Figure 14. The average speed for the 6 segments multi-mode in pipe locomotion under different gaits shown in Table 1.

## 5. Conclusion and discussion

In this paper, a multi-mode underwater locomotion robot based on earthworm is designed and prototyped. The robot not only can crawl like the earthworm in the pipe, but also can swim in the water. The main results of our work are presented as follows:

1. The robot design and fabrication: according to the earthworm's multi-segment and muscle characteristics, we fabricate the acrawl module, which contains six earthworm-like segments. To prevent the water to destroy the servo motor, several moulds are designed to fabricate silicone skins. To increase the locomotion ability, the swimming module is designed, which include the left and right propellers to drive the robot moves on the horizontal plane, and front and back propellers to drive the robot moves on the vertical plane.

2. The controller design and experiments: the computer system generates signals and sents to the Arduino control board. After being handled, the signals are sent to the servo motor control board to drive the six earthworm-like segments, and the DC motor drivers to drive the four propellers. For the experiment, we test the robot on the crawling mode, the swimming mode, and mode change from the crawling mode to the swimming mode.

3. The dynamic equations corresponding to the horizontal plane and the vertical plane are established based on the submarine's dynamic equations to predict the robot's locomotion. Especially, the robot can move with circular locomotion on the horizontal plane, linear locomotion on the horizontal and vertical plane. Furthermore, the CFD software is used to simulate the three basic locomotions and compared them with ODE45 numerical simulation based on the dynamic equations. And further analysis for the three basic locomotion is studied with the dynamic equations.

4. The kinematic model for in-pipe locomotion is cited to analyze the velocities for the six different gaits.

The earthworm-like robot can be used for many environments, however, the movement efficiency is lower. To use for reference to earthworms' advantages and improve the movement efficiency, the swimming mode is designed to combine with the earthworm-like segments. The results review that the multi-mode robot not only can crawl like the earthworm in the pipe but also can swim in the water.

However, the robot can't autonomously explore the environment. So the content of the paper is the first step in the discovery of the multi-mode underwater robot. In the future, the sensors will be installed to feedback the information of the environment to independently plan the path and avoid obstacles.